# Context-Aware Sequence-to-Sequence Models for Conversational Systems

Silje Christensen, Simen Johnsrud, Massimiliano Ruocco and Heri Ramampiaro
Norwegian University of Science and Technology
NTNU, Trondheim N-7491, Norway
{ruocco,heri}@ntnu.no

## ABSTRACT

This work proposes a novel approach based on sequence-to-sequence (seq2seq) models for context-aware conversational systems. Existing seq2seq models have been shown to be good for generating natural responses in a data-driven conversational system. However, they still lack mechanisms to incorporate previous conversation turns. We investigate RNN-based methods that efficiently integrate previous turns as a context for generating responses. Overall, our experimental results based on human judgment demonstrate the feasibility and effectiveness of the proposed approach.

## 1 INTRODUCTION

The Sequence-to-Sequence (seq2seq) model is one of the most successful architectures for generating responses in a conversational system, see, e.g., [13, 14]. It is a supervised and data-driven approach, where an utterance is encoded through a Recurrent Neural Network (RNN) based encoding mechanism and used as a starting point for the generation of the response through an RNN-based decoder. The main weakness of this approach and its variants is that they do not handle information from preceding dialogue turns in a conversation, but simply return the highest ranked answer to the current question. This can, for example, be problematic with questions such as "*How do I do that?*", which need a context to make sense. A sensible solution is to mimic human conversations. To achieve this, however, we have to look at prior turns to help the conversational agent make use of the information from these turns.

Inspired by the work in [10, 12], we propose a new approach that can incorporate previous dialogue utterances (context) in the original seq2seq-based architecture. Moreover, we analyse the impact of using different unit cells in the encoder-decoder architecture for the response generation task. In particular, we study the performance of the GridLSTM [5], which has the distinguishing feature that it does not compress the input vector into a single vector.

At the outset of this work the main goals were: (1) to study the impacts of the use of different RNN cells in an Encoder-Decoder model on the quality of the outputs, (2) to develop a seq2seq-based model that is able to keep track of previous questions and responses, in order to catch the conversation context better. We proceeded in an iterative manner: First, we implemented an RNN encoder-decoder model using LSTM cells as proposed in [14] with an attention mechanism [1] as baseline, and compared this model to the models that we propose in this work. Since automatic evaluation of conversational agents is still an open problem, and that existing automatic evaluation methods, such as BLEU [9], are most suitable for machine translation-related problems, we follow [7, 11] and evaluate them by conducting an extensive human evaluation. Thus, we compared methods applying different RNN cells in a conversational agent, and used the results from this comparison to identify further required changes. Specifically, we change the preprocessing procedure, handle the internal state during training, and modify the decoder. The effects of these changes are then compared against each other. To evaluate the different conversational agents we use our datasets to extract test questions or create fictive conversations (see Section 4.1). Finally, the agents' replies are used in questionnaires for a 5-scale human evaluation method.

The main contributions of this work can be summarised as follows: First, in order to improve the ability of the model to use information from previous turns, we propose a stateful model as a result of the adjustments in the encoder-decoder architecture. To the best of our knowledge, this work is the first application of seq2seq for conversational agents that is trained in a stateful manner. Second, we compare GRU, LSTM, and GridLSTM cells in the same architecture. This is in itself a new approach to develop a neural network-based architecture for conversational systems. Third, we propose a novel approach for eliminating out-of-vocabulary (OOV) words from the dataset to further improve the accuracy of the conversations. Overall, the approach proposed in this work is generic, and is applicable in many applications using conversational systems, including conversational IR systems.

## 2 RELATED WORK

Research on conversational systems can be divided into two categories: (1) *retrieval*-based, e.g., [3, 6]; and (2) *generative*-based, e.g., [7, 8, 10, 12, 14]. Systems in category (1) are those using a repository of predefined responses, while systems in category (2) consist of those that are often based on machine translation methods. This work mainly falls in category (2).

Ritter et al. [10] proposed one of the first approaches to treat the response generation problem as a statistical machine translation problem, but the approach was not sensitive to the context of the conversation. The work by Sordoni et al. [12] had, on the other hand, a better success in making the translation model context-sensitive by incorporating previous turns using RNN-based Language Model. Further, the effort on seq2seq model for machine translation by Sutskever et al. [13] has inspired a set of approaches in the area of generative dialogue systems, including [8] and [14]. Note, however, that working with a dialogue system is generally less challenging than addressing the translation problem because of the availability of context and well-defined evaluation methods, e.g., BLEU [9]. In line with this, Sutskever et al. [13] proposed an RNN Encoder-Decoder approach to solve the machine translation problem. Bahdanau et al. [1] addressed weakness of the previous architectures in handling long sentences by adding a so-called attention mechanism to the Encoder-Decoder approach, thus making it possible for the decoder to decide which parts of the source sentence to pay attention to. Vinyals and Le [14] employed the architecture described in [13] for building a model to generate responses in conversational systems, with the aim to show the ability to produce natural conversations. However, their model did not take the context of a conversation into account, thus resulting in generating possibly contradicting replies within a dialogue. Finally, Shang et al. [11] proposed a Neural Responding Machine, employing the Encoder-Decoder framework using GRU-cells to address the response generation problem. Unlike the previously mentioned approaches, they focused on Short-Text Conversation (i.e., two turns) problem instead of translation and using a three types of encoding scheme. Similar to our work, their evaluation is done with human judgment.

## 3 PROPOSED CONVERSATIONAL MODELS

### 3.1 Baseline models

To be able to evaluate our proposed models, we decided to implement a baseline by integrating the original seq2seq architecture/model proposed by Cho et al. [2] with an attention mechanism [1]. Doing this allows for each new word by the model to focus on specific parts of the original text. In brief, the (encoder-decoder) model works as follows. The encoder is fed into the RNN with the embedded input tokens, which outputs an hidden state $h_i$ each time, mainly keeping all valuable information of the sequence. These hidden states $h_i$ are also part of the inputs for the attention model and it is generally computed in RNN as follows: $h_t = f(\mathbf{x}_t, h_{t-1})$, where $f$ is a nonlinear activation function.

The decoder generates the sequence of tokens representing the response to the turn from the input. This is similar to the RNN encoder. The main difference is given by the additional input $c_i$ from the attention model, and is mainly used to infer the attention from the source inputs from the encoder in the prediction of the next token. In other words, to predict the next token $y_i$ (via softmax), each unit of the decoder uses the context vector $c_i$, the previous hidden state $h'_{i-1}$, and the previous output $y_{i-1}$ as $P(y_i|y_{i-1}, y_{i-2}, ..., y_1, \mathbf{x}) = g(y_{i-1}, h'_i, c_i)$, where the hidden state of the decoder is computed by $h'_i = f(h'_{i-1}, y_{i-1}, c_i)$. $c_i$ is calculated for each target word $y_i$ as the weighted average of the hidden states coming from the encoders.

**LSTM and GRU:** To avoid the well-known vanishing gradients problem for long sequences, we utilize the GRU (Gated Recurrent Unit) [2] and LSTM ( Long-Short Term Memory) [4] as unit cells of both the encoder and the decoder RNNs. LSTM and GRU are both specifically interesting due to their ability to learn long term dependencies, with GRU being less complex than LSTM.

**Grid LSTM-based Seq2seq**: Grid LSTM [5] is an extension of LSTM by arranging LSTM cells in a multidimensional grid. This makes it applicable to vectors, sequences or higher dimensional data. Another feature of Grid LSTM is that the cells in the grid can communicate with each other across different layers. Motivated by the challenging task of representing all of the information in a sentence in one vector, we propose to use the Grid LSTM cells in the seq2seq model. The aforementioned attention mechanism relieves some of the pressure on the encoder, which the Grid LSTM further improves as a result of the structure of the Grid LSTM block. Moreover, with Grid LSTM the input does not need to be compressed into one vector. Instead, each word in the input sequence is projected on one side of the grid. Thus, the model can scan a source sentence repeatedly. In this work, we study how this affect the quality of the output from a conversational agent.

### 3.2 Context-based Models

We propose three RNN-based architectures for conversational agents, called Stateful Model, Stateful-Decoder, Context-Prepro. They extend the *Neural Conversational Model*, i.e., Sequence-to-Sequence (seq2seq) model [14], and the *encoder-decoder* architecture [13].

**Stateful Model:** Previous architectures have not considered information from preceding turns in a conversation, but simply returns the best predicted answer to the current question. To mimic a human conversation, however, it is crucial for the conversational agent to look at prior turns for it to be able to reuse this information in its answer, independent of whether the most recent question might not contain any information about the topic or not. The idea behind our Stateful model is to use the previously generated state in the RNN decoder as the initial state for the next turn. In this model, the encoder is similar to the encoder in the LSTM-based seq2seq baseline, but we had to make some adaptations in the RNN decoder. Note that for a non-stateful model, it is straightforward to train the model using batches, because the training steps are independent of each other. However, this independence can result in every batch to obtain multiple decoder states. For the Stateful model, on the other hand, we need to connect the training steps together, because we want to pass the decoder state to the next input. The reason for this is that the previous state should have an impact on the next sentence in a conversation. To the best of our knowledge, this is the first application of an RNN model in a conversational agent that is trained in a stateful manner.

**Stateful-Decoder:** With our context models, we only output the first sentence from the decoder to avoid confusion. This is also because we cannot guarantee the quality of previous responses. A stateful decoder will, just as for the stateful training, pass on the decoder states during the conversation. The Stateful-Decoder model is identical to the LSTM baseline with one bucket during training, but during decoding, it is identical to the Stateful model.

**Context-Prepro:** For a fair comparison, we also propose a slight modification of the LSTM-based seq2seq baseline, with which the difference is that the input consist of the previous response and the current question. To do this, we concatenate these sentences before sending the resulting sentence into the encoder.

## 4 EXPERIMENTAL SETTING

### 4.1 Datasets and data preparation

**Datasets:** To evaluate the our models, we used *UDC* and *Open Subtitles* datasets. UDC is a closed domain dataset based on the Ubuntu Chat Logs (https://irclogs.ubuntu.com). It is a forum used to discuss technical issues concerning the Ubuntu operative system, consisting of 8.6$M$ turns built of 168.7$M$ words. We chose this dataset because of its interesting characteristics. It is noisy (e.g., a response can be only a URL), and a user may type several consecutive responses (i.e., long turns). OpenSubtitles (http://www.opensubtitles.org) consists of conversations from movie manuscripts. It can be viewed an open domain dataset since there are no restrictions on topics in movie manuscripts. Thus, it is interesting to study how a conversational agent trained on such a dataset would do in a chit-chatting task. We extracted 2.7$M$ sentences, built of 19$M$ words. Note that because OpenSubtitles is based on movie subtitles, we can assume some grammatical regularity, but the manuscripts are not provided with any information about which actor who says what. Hence, we do not know if a given sentence is a reply to a previous sentence, or if one person speaks more than once. Therefore, it is particularly hard to extract good question and response pairs.

**General data preparation:** After extracting the question-response pairs from the datasets, we decided to reduce the vocabulary size to decrease the training and decoding complexity by removing the least frequent out-of-vocabulary (OOV) words. To avoid possible issues, such as unknown token output and destruction of sentence structure, we applied our own strategy to replace OOV words. First, we replaced every OOV word with the most similar word in the vocabulary based on both morphology and semantic, by first replacing special tokens, e.g., urls and directory paths, and correcting misspelling errors, and then, after analyzing the tokens distributions, we ended up keeping a set of tokens representing around 96% of the dataset. This reduced the number of unique words from 2.4M to 1.3M for the UDC dataset and from 128K to 7K for the OpenSubtitles dataset. Second, we replaced the tokens in OOV with the most similar words in the vocabulary with respect to the embedding representations. Here, we used FastText (https://github.com/facebookresearch/fastText) to generate the word embeddings.

**Model-specific data preprocessing:** Both the Stateful and the Context-Prepro model required further and special data preparation. For the Stateful Model, the question and response must be listed chronologically. Hence, the order of the training pairs within a conversation must stay unshuffled, while the conversation itself can be shuffled among the other conversations. Further, no turns in the conversation can be longer than a predefined bucket size. Therefore, we removed all conversations with turns whose size is longer than this. For Context-Prepro, to include the context in the training data, the previous answer was added in front of the next training input, which required an expansion of the bucket sizes. Due to the trade-off between the amount of information we should concatenate as input to the model and the training complexity, we decided to limit the context to the previous answer only.

### 4.2 Experimental Plan

We carried out our experiments in three main parts as follows:
**Part 1** concerns testing the RNN Cells, and studying the impacts of applying different them in an encoder-decoder model on the quality of the outputs. In addition, we compare GridLSTM seq2seq-model (`S2S_GridLSTM`) against the LSTM and GRU seq2seq models (`S2S_LSTM` and `S2S_GRU`) focusing on how they handle context in a sentence.
**Part 2** deals with exploring the context approaches, i.e., evaluation of a conversational agent's ability to capture the context of a conversation based information from previous turns. Here, we compare the stateful (`S2S_Stateful`) and `Context-Prepro` models against the baseline, i.e., `S2S_LSTM` trained on the original preprocessed UDC dataset.
**Part 3** focuses on increasing the external validity of our models. This means we evaluate the whole approach, including the Stateful Decoder `S2S_Stateful-Decoder`. We study how the `S2S_GridLSTM` model handles chit-chatting compared to the `S2S_LSTM` baseline, using OpenSubtitles. Because a single movie manuscript usually is longer than the normal UDC conversation, and the context changes several times due to different scenes in a movie, we do not train the context models on this dataset. As a compromise, we embed the `S2S_Stateful-Decoder` to the baseline, to see if the decoder can catch the context of a whole conversation, even though the training procedure did not follow the stateful approach.

### 4.3 Evaluation Metrics

Due to the still lack of ground truths, we conducted a 5-scale human evaluation (similar to [11]). This means that, for each question or conversation, the evaluator was asked to rate the answer(s) from the different conversational agents, with a score from 1 to 5. This rating includes rating both the *grammar*, i.e., the sentence structure and the spelling, and *content*, i.e., the quality of the response by an agent to a question, as well as, how well the agent remembers previous utterances as in a normal human conversation.

We split the experiments into three groups distributed in two questionnaires. The first two groups were evaluated with the first questionnaire, while group 3 was evaluated in the second one. For group 1 and group 2, we extracted the questions from the corpus' test set. Due to the technical content, the evaluators were mainly 23 and 27 years old ICT students. We also supplied some of the questions with additional information and the evaluator was asked to rate single questions. We chose to include the responses from the test set in the questionnaire, to get an idea of how well the models performed compared to the content in the dataset. In this part, the evaluators will rate whole conversations from the conversational agent, where the questions may differ as the conversations evolve. For group 3, we involved 50 persons between 20 and 64, with no strict requirement on technical knowledge. Due to the issue related to the OpenSubtitles dataset, we did not get questions from the test set for the questionnaire. Instead, we defined seven different topics, and asked the questions that we found suitable within the topics.

## 5 RESULTS AND DISCUSSION

**Part 1:** In this part of the experiments, we trained S2S_GridLSTM, S2S_LSTM and S2S_GRU model on the UDC dataset. Table 1 presents the results of the human evaluation based on 12 randomly selected questions. As shown, the actual response from the dataset is superior to the implemented models, which was not a surprise. It is, however, interesting that the test set did not receive an even better score. Moreover, the fact that the LSTM model obtained better results than the S2S_GRU model totally made sense, given that the S2S_GRU is a *simplification* of the LSTM cell. The difference between the S2S_GridLSTM and S2S_LSTM's total score is only 0.02. Hence, these two models were virtually equivalent. S2S_GRU, on the other hand, obtained the lowest score, so we decided to discard this model, although it had a lowest training time.

|  | Grammar | Content | Total score |
|---|---|---|---|
| **S2S_GridLSTM** | **3.59** | **3.00** | **3.29** |
| **S2S_LSTM** | **3.61** | **3.01** | **3.31** |
| **S2S_GRU** | 3.45 | 2.91 | 3.18 |
| *Ref. Dataset* | *3.86* | *3.69* | *3.77* |

Table 1: Results from Part 1 of the evaluation, with single questions from the UDC

|  | Grammar | Content | Total score |
|---|---|---|---|
| **S2S_Stateful** | **3.80** | **2.71** | **3.25** |
| **S2S_LSTM** | **3.78** | 2.38 | 3.08 |
| **Context-Prepro** | 3.75 | 2.08 | 2.92 |
| **Dataset** | 4.23 | 3.98 | 4.1 |

Table 2: Results from Part 2 of the evaluation considering the UDC conversations

|  | Grammar | Content | Total score |
|---|---|---|---|
| **S2S_GridLSTM** | **4.14** | **3.26** | **3.70** |
| **S2S_Stateful-Decoder** | 3.97 | 2.67 | 3.32 |
| **One-Bucket** | 3.80 | 2.78 | 3.29 |
| **S2S_LSTM** | 3.91 | 2.67 | 3.29 |

Table 3: Results from Part 3 of the evaluation using OpenSubtitles

**Part 2:** The results from the human evaluation from this part of the experiment are shown in Table 2. Here, the agents tested in group 2 score worse than the models in group 1. Note, however, that this part focused on a greater problem, since it did not only consider single questions, but an entire conversation. As shown, the S2S_Stateful model received the highest score among the different models, while the baseline outperformed the Context-Prepro model. This difference indicates that the S2S_Stateful model manages to capture information from previous turns better than non-stateful models, regarding the content, and then passing the previous state to the next turn helped the agent when interpreting the conversation. The system's response reflected the context, even though the human question did not contain any information about the topic. As in Part 1, the test set is again superior, and this time with a greater margin. The increased score for the test set makes sense since longer conversations with reasonable content are more difficult than single sentences. Another interesting observation is that the LSTM model in group 1 had a score of 3.31, whereas the identical model used for group 2 got a score of 3.08. All this means that it is hard for conversational agents to respond properly several times in a row and to substitute a human in a conversation with several turns.

**Part 3:** Recall that in this part, we focused on evaluating our whole approach. The results from this evaluation are shown in Table 3. Here, S2S_GridLSTM model got an average score of 3.70 and a content score of 3.26, which made it the best model in this experiment. The S2S_LSTM baseline had an average score of 3.29. These results indicates that the Grid LSTM was better than the standard LSTM cell. Note that using the S2S_Stateful-Decoder on a non-stateful model did not yield any significant improvements. In fact, if we look at the content score, the S2S_LSTM model trained on one bucket performed better than the equivalent model using the stateful decoder. Unlike Part 1, Part 3 showed a significant difference between the S2S_GridLSTM model and the S2S_LSTM baseline. A content score of approx. 0.5 points better than the next best model shows that the use of GridLSTM cells improves the results for the chit-chatting task. The minimal effect on the results with the S2S_Stateful-Decoder in Part 3, and the fact that the S2S_Stateful model was superior in Part 2 indicates that the S2S_Stateful-decoder itself has little effect on non-stateful models. Another observation is that the models with a single bucket tend to generate longer responses compared to both the S2S_LSTM and S2S_GridLSTM model.

## 6 CONCLUSION

In this work, we have studied different RNN-based models that incorporate context information to mimic human conversation in conversational systems. We have proposed a stateful model which extends the seq2seq model by improving its ability to use and remember the information from previous turns. The responses generated from the model have been evaluated with human evaluation. The results from this evaluation have shown that the proposed model is better than the baselines, with respect to the content of the responses. We have also learned how the choice of the unit cell in the RNN-based encoder and decoder affect the quality of the output, showing in particular how the use of the GridLSTM cell can increase the content quality of responses in casual conversations and small talks.

## REFERENCES

[1] Dzmitry Bahdanau, Kyunghyun Cho, and Yoshua Bengio. 2014. Neural machine translation by jointly learning to align and translate. *CoRR* abs/1409.0473 (2014). http://arxiv.org/abs/1409.0473
[2] Kyunghyun Cho, Bart Van Merriënboer, Caglar Gulcehre, Dzmitry Bahdanau, Fethi Bougares, Holger Schwenk, and Yoshua Bengio. 2014. Learning phrase representations using RNN encoder-decoder for statistical machine translation. *CoRR* abs/1406.1078 (2014). http://arxiv.org/abs/1406.1078
[3] Jiahui Guo, Bin Yue, Guandong Xu, Zhenglu Yang, and Jin-Mao Wei. 2017. An Enhanced Convolutional Neural Network Model for Answer Selection. In *Proc. of WWW 2017 Companion*.
[4] Sepp Hochreiter and Jürgen Schmidhuber. 1997. Long short-term memory. *Neural computation* 9, 8 (1997), 1735–1780.
[5] Nal Kalchbrenner, Ivo Danihelka, and Alex Graves. 2015. Grid long short-term memory. *CoRR* abs/1507.01526 (2015). http://arxiv.org/abs/1507.01526
[6] Vineet Kumar and Sachindra Joshi. 2017. Incomplete Follow-up Question Resolution Using Retrieval Based Sequence to Sequence Learning. In *Proc. of SIGIR 2017*. ACM, 705–714.
[7] Jiwei Li, Michel Galley, Chris Brockett, Georgios P. Spithourakis, Jianfeng Gao, and William B. Dolan. 2016. A Persona-Based Neural Conversation Model. In *Proc. of ACL 2016*. ACL.
[8] Chuwei Luo, Wenjie Li, Qiang Chen, editor="Jose Joemon M He, Yanxiang", Claudia Hauff, Ismail Sengor Altıngovde, Dawei Song, Dyaa Albakour, Stuart Watt, and John Tait. 2017. *A Part-of-Speech Enhanced Neural Conversation Model*. Springer, 173–185.
[9] Kishore Papineni, Salim Roukos, Todd Ward, and Wei-Jing Zhu. 2002. BLEU: a method for automatic evaluation of machine translation. In *Proc. of ACL (2002)*. ACL, 311–318.
[10] Alan Ritter, Colin Cherry, and William B Dolan. 2011. Data-driven response generation in social media. In *EMNLP 2011*. ACL, 583–593.
[11] Lifeng Shang, Zhengdong Lu, and Hang Li. 2015. Neural responding machine for short-text conversation. *CoRR* abs/1503.02364 (2015). http://arxiv.org/abs/1503.02364
[12] Alessandro Sordoni, Michel Galley, Michael Auli, Chris Brockett, Yangfeng Ji, Margaret Mitchell, Jian-Yun Nie, Jianfeng Gao, and Bill Dolan. 2015. A neural

network approach to context-sensitive generation of conversational responses. *CoRR* abs/1506.06714 (2015). http://arxiv.org/abs/1506.06714
[13] Ilya Sutskever, Oriol Vinyals, and Quoc V Le. 2014. Sequence to sequence learning with neural networks. In *NIPS 2014*. 3104–3112.
[14] Oriol Vinyals and Quoc V. Le. 2015. A Neural Conversational Model. *CoRR* abs/1506.05869 (2015). http://arxiv.org/abs/1506.05869